\documentclass[]{interact}

\usepackage{subcaption}
\usepackage{hyperref}
\usepackage[font=small,labelfont=bf]{caption}
\usepackage[section]{placeins}

\usepackage{natbib}
\bibpunct[, ]{(}{)}{;}{a}{}{,}
\renewcommand\bibfont{\fontsize{10}{12}\selectfont}

\begin{document}

\title{There Is a Digital Art History}

\author{
\name{Leonardo Impett\textsuperscript{a*} and Fabian Offert\textsuperscript{b*}}
\affil{\textsuperscript{a}Cambridge University, United Kingdom; \textsuperscript{b}University of California, Santa Barbara, U.S.A.
\newline
* Equal contributions } }

\maketitle

\begin{abstract}
In this paper, we revisit Johanna Drucker's question, ``Is there a digital art history?'' -- posed exactly a decade ago -- in the light of the emergence of large-scale, transformer-based vision models. While more traditional types of neural networks have been part of digital art history for some time, and transformer-based vision models have been part of digital humanities projects at least since 2020, their epistemic implications and methodological affordances have not yet been systematically analyzed. We focus our analysis on two main aspects that, together, seem to suggest a coming paradigm shift towards a ``digital'' art history in Drucker's sense. On the one hand, we argue that the visual-cultural repertoire newly encoded in large scale vision models has an outsized effect on digital art history. The inclusion of significant numbers of non-photographic images allows for the extraction and automation of different forms of visual logics, including non-perspectival, non-figurative, and non-representational logics. Large-scale vision models have ``seen'' large parts of the Western visual canon mediated by Net visual culture, and they continuously solidify and concretize this canon through their already widespread application in all aspects of digital life. On the other hand, based on two technical case studies of utilizing a contemporary large-scale visual model to investigate basic questions from the fields of art history and urbanism, we suggest that such systems require a new critical methodology that takes into account the epistemic entanglement of a model and its applications. This new methodology reads its corpora through a neural model's training data, and vice versa: the visual ideologies of research datasets and training datasets become entangled.
\end{abstract}

\renewcommand{\abstractname}{SOFTWARE}
\begin{abstract}
We make the tools developed for both experiments in this study freely available on the Web so
that our results can be reproduced on demand: CLIP-MAP (\url{https://leoimpett.github.io/clip-map/}), 2D-CLIP (\url{https://leoimpett.github.io/2dclip/}).
\end{abstract}

\section{Introduction}\label{introduction}

In 2013, Johanna Drucker asked: \emph{Is There a `Digital' Art History?} Her provocatively-titled article suggests an important difference between ``digitized'' and ``digital'' art history. While the former --- the ``making digital'' of visual culture --- describes a somewhat successful set of practices, the latter is more nebulous. There has been ``no research breakthrough''\footnote{Johanna Drucker, ``Is There a `Digital' Art History?'', \emph{Visual Resources} 29, no. 1-2: 5.} so far, Drucker argues, and we have yet to witness ``a convincing demonstration that digital methods change the way we understand the objects of our inquiry''.\footnote{Drucker, ``Is There a `Digital' Art History?'': 6. Claire Bishop, five years later, doubles down on Drucker's claims: ``Theoretical problems are steamrollered flat by the weight of data.'' Claire Bishop, ``Against Digital Art History,'' \emph{International Journal for Digital Art History} 3 (2018): 125.} While we can always have more convenience through digitization, a digital art history that enables new kinds of questions has yet to emerge.

Drucker's argument, of course, mirrors the fundamental methodological question of the digital humanities in general --- what is the ``surplus value''\footnote{The recurring accusation that the digital humanities are a ``neoliberal'' field, or at least complicit in the radical transformation of the university under late capitalism, hinges on this foundational search for the ``surplus value'' of the digital. Traditional humanistic methods are already saying everything there is to say about culture -- or so the conservative argument goes -- such that any additional mode of analysis becomes extractive, i.e. destructive.} of the digital? One answer, epitomized in the notion of ``distant reading''\footnote{Franco Moretti, Distant Reading (New York, NY: Verso Books, 2013).}/''distant viewing''\footnote{Taylor Arnold and Lauren Tilton, ``Distant Viewing: Analyzing Large Visual Corpora,'' \emph{Digital Scholarship in the Humanities}, 2019.} is scale. But the rare works of digital art history that have focused on intrinsically large-scale problems -- Diana Greenwald on the economic history of 19th century Western art production,\footnote{Diana Seave Greenwald, \emph{Painting by Numbers: Data-driven Histories of Nineteenth-century Art} (Princeton University Press, 2021).} or Matthew Lincoln on networks of early modern Dutch and Flemish printmaking\footnote{Matthew David Lincoln, \emph{Modeling the Network of Dutch and Flemish Print Production, 1550-1750}, PhD dissertation (University of Maryland, College Park, 2016).} -- have tended to do so outside the realm of the (digital) image. In the past decade, \emph{visual} digital art history has often been arrested by a very basic question, which turned out to be a significant technical challenge: what is \emph{in} an image?

The primary reason for this divergence of digital art history and the digital humanities at large lies in what we could call the Laocoön problem of computation. In short, the affordances of images are different from the affordances of text. Images and text, in fact, are almost diametrically opposed in the digital realm. Not only is there no equivalent to the discrete ``tokens'' of text-based digital approaches in the visual domain, but there is also no commonly-agreed hierarchy of elements (the paragraph, the sentence, the word). Sentences end with a period. But where do image-objects end, exactly? Most texts have a clearly defined vocabulary -- some number of words or subwords, with a reasonable upper bound. In images, potential ``vocabularies'' are ambiguous, infinite, and not tied to any common superset. And even if images share a common -- for instance iconographic -- visual vocabulary, no instance of a ``word'' is like the other.\footnote{Even bigger challenges emerge from abstract art, sculpture, or performance art. For the latter, see Miguel Escobar Varela, \emph{Theater as Data: Computational Journeys into Theater Research} (University of Michigan Press: 2020).} The ingenuity of Lev Manovich's pioneering work on image sets qua \emph{style spaces}\footnote{Lev Manovich, ``Style Space: How to Compare Image Sets and Follow Their Evolution,'' 2011, URL: \url{http://manovich.net/content/04-projects/073-style-space/70_article_2011.pdf} (accessed 7/15/2023). See also Lev Manovich, \emph{Cultural Analytics} (MIT Press: 2020).} is to sidestep this question completely: digital images are pixels, and pixels have measurable properties like color, brightness, and entropy. At the same time, this approach limits our inquiry to the kind of phenomena measurable at the pixel level. The challenges for digital methods in art history, then, seem different -- and perhaps more fundamental\footnote{For a history of computer vision that touches upon some of these differences, see James Dobson, \emph{The Birth of Computer Vision} (University of Minnesota Press, 2023).} -- from those in other branches of the digital humanities. Johanna Drucker's question still stands, ten years after it was first proposed: has there been a ``digital'' art history?

In this essay, we suggest that a ``digital'' art history is, at the very least, on the horizon. It emerges from a necessary entanglement of \emph{data analysis} and \emph{model critique} in contemporary multimodal models, machine learning systems that are trained on both textual and visual data. In fact, we argue, multimodal models can \emph{only} contribute to art historical scholarship if this entanglement of analysis and critique is taken seriously. There cannot exist a visual analysis of images using multimodal models that is not also, and at the same time, a critique of the conceptual space inherent in the model. Tool and data, in the age of multimodal models, exist in a reciprocal relationship: looking at data with multimodal models means looking at multimodal models with data. Given the current trajectory of machine learning research, this methodological shift is here to stay.

\section{Digital Art History as Object Recognition}\label{digital-art-history-as-object-recognition}

\begin{figure}[h]
\centering
\includegraphics[width=\linewidth]{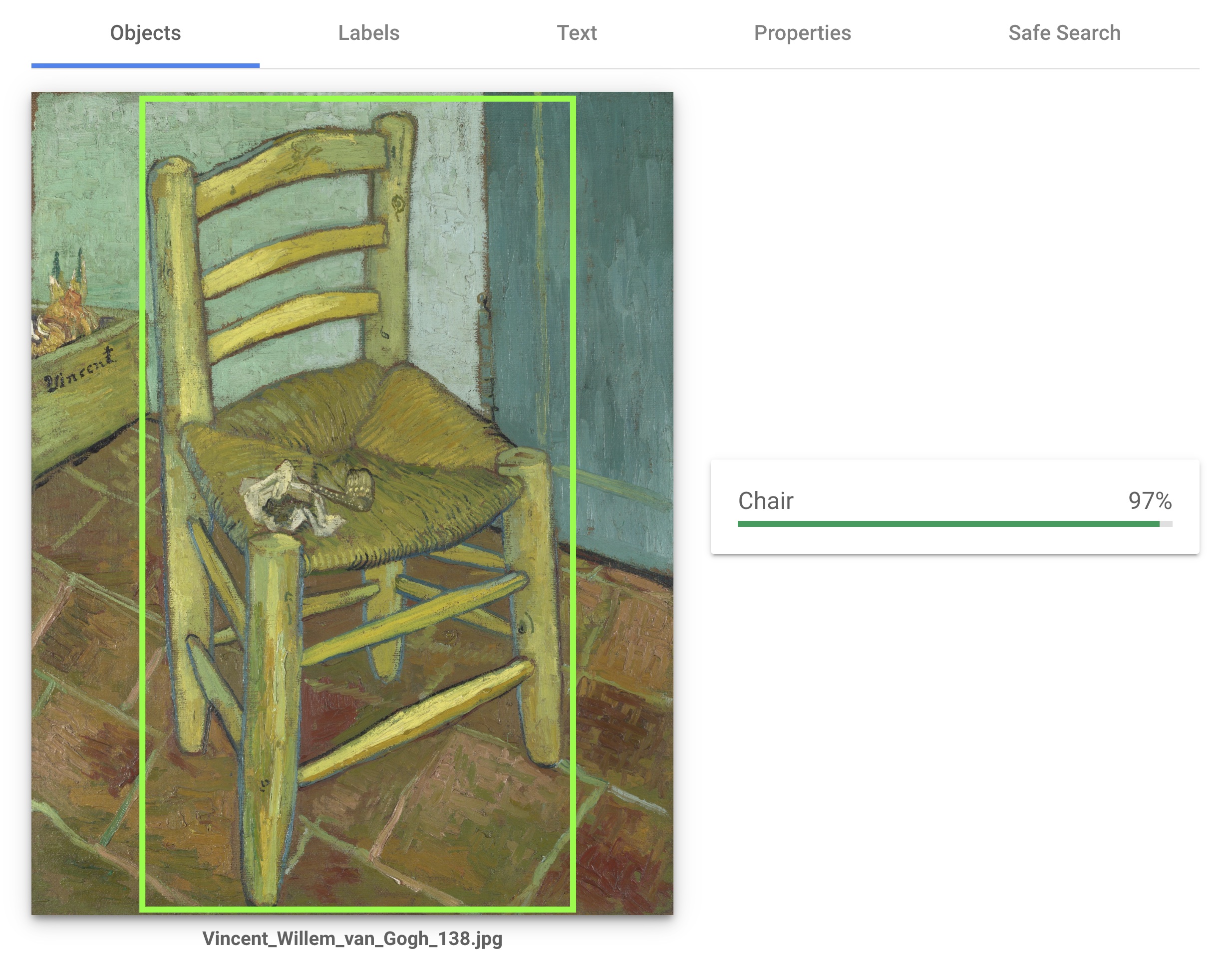}
\caption{Van Gogh's chair detected by computer vision. We use the Google cloud vision API as an example of a widely used object recognition model.}
\label{fig1}
\end{figure}

In a 2012 interview with Chris Wood, art historian Horst Bredekamp was asked what the impact of machines that can read images will be on the history of art. He answered: ``No, it's impossible for machines. Specialists said recently that even in a thousand years a computer will not be able to recognize the chair painted by Vincent van Gogh as a chair. Computers would need bodies, as the discussion on Körperschema has shown\ldots{} That is one of the consequences of embodiment philosophy {[}\ldots{]}.''\footnote{Christopher Wood and Horst Bredekamp, ``Iconoclasts and Iconophiles: Horst Bredekamp in Conversation with Christopher Wood,'' \emph{Art Bulletin}, December 2012: 526f.} As figure \ref{fig1} makes clear, Bredekamp was ill-advised. Since 2015, deep convolutional neural networks in particular have led to great leaps in the power of machines to recognize objects within images. A number of scholars have turned this new capacity for detection onto art-historical problems. In the realm of ``thing'' detection, the study of iconographic patterns\footnote{Peter Bell and Leonardo Impett, ``The Choreography of the Annunciation through a Computational Eye,'' \emph{Histoire de l'Art} 34 no. 87, 2021: 1-6.} has benefitted from the robustness of pre-trained classifiers\footnote{Artificial neural networks trained on standard datasets, most prominently ImageNet} vis-a-vis figurative works in particular. Pose detection has facilitated the quantitative study of art-historical concepts related to gesture, for instance in Impett's and Moretti's study of Aby Warburg's \emph{Pathosformel}.\footnote{Leonardo Impett and Franco Moretti, ``Totentanz. Operationalizing Aby Warburg's \emph{Pathosformeln},'' Stanford Literary Lab Pamphlet 16, 2017. See also Peter Bell and Leonardo Impett, ``Ikonographie und Interaktion. Computergestützte Analyse von Posen in Bildern der Heilsgeschichte,'' \emph{Das Mittelalter} 24 no. 1, 2019: 31-53.} Finally, face recognition has allowed scholars to study the appearance and reappearance of historical and fictional ``characters'' at scale.\footnote{Prathmesh Madhu, Ronak Kosti, Lara Mührenberg, Peter Bell, Andreas Maier, and Vincent Christlein, ``Recognizing Characters in Art History Using Deep Learning,'' \emph{Proceedings of the 1st Workshop on Structuring and Understanding of Multimedia Heritage Contents}, 2019: 15-22.}

To be clear: we group all of these approaches\footnote{A recent overview and critique of the state of the art of object recognition can be found in Amanda Wasielewski, \emph{Computational Formalism. Art History and Machine Learning} (MIT Press: 2023).} together under ``object recognition'' for two reasons. On the one hand, they all utilize a class of machine learning models -- deep convolutional neural networks -- that are more or less directly derived from object classification or classification models. On the other hand, and more significantly: all of these approaches attempt to point to (by pixel-wise segmentation, bounding boxes, or classification) semantically unambiguous \emph{things} (``objects'') depicted in images. These models are most often trained on human annotations, and it is assumed that a human observer can verify the model's predictions -- giving some route into identifying, if not explaining, the errors or biases in a model. This form of object recognition (if we are indeed happy to call human limbs and faces ``objects'') is thus certainly useful. But what modes of understanding, what forms of knowledge, does it afford? What model of cognition sits underneath an image recognition model?

Vision models for object-related tasks are trained on a predetermined set of object categories: handwritten arabic numerals in MNIST, or a combination of natural species (bird, cat, marmoset) and manufactured objects (automobile, television, Polaroid camera) in CIFAR-10 or ImageNet. To learn to classify the MNIST dataset, a network doesn't need to know that ``2'' falls between ``1'' and ``3'', or indeed that it has anything to do with mathematics or writing. Instead, for neural networks, object/image categories are arbitrary labels that are literally discarded when training, and reattached post hoc, merely for the convenience of the human user. Object detection, in other words, always relies on rigid ontologies which are in themselves meaningless to the computer. The forms of understanding it affords are still necessarily anagraphic, tied to preselected slices of the conceptual, not the physical world. It is this anagraphic quality that also fuels the myth of the objectivity of object recognition. Where there is nothing beyond representation, objectivity, which itself emerges from historical forms of visual representation,\footnote{Lorraine Daston and Peter Galison. \emph{Objectivity} (Princeton University Press: 2021).} becomes the assumed norm.

The ontologies of CIFAR or ImageNet also betray a phantom afterlife of GOFAI (``Good Old-Fashioned AI''), the symbolic, rules-based approach to AI popular until the 1990s. Rules-based systems were good at solving problems like chess, but unable to contribute much to problems of perceiving their environment.\footnote{Hans Moravec, \emph{Mind Children: The Future of Robot and Human Intelligence} (Harvard University Press: 1988).} As Rodney Brookes put it in \emph{Elephants Don't Play Chess}: ``it is necessary to have its {[}the AI system's{]} representations grounded in the physical world\ldots{} once this commitment is made, the need for traditional symbolic representations soon fades entirely''.\footnote{Rodney A. Brooks, ``Elephants Don't Play Chess,'' \emph{Robotics and Autonomous Systems} 6 no. 1-2: 5.} Hubert Dreyfus later famously concluded\footnote{Hubert L. Dreyfus, ``Why Heideggerian AI Failed and How Fixing It Would Require Making It More Heideggerian,'' \emph{Philosophical Psychology} 20 no. 2: 247--68.} that GOFAI, even where it took an explicit ``Heideggerian'' stance such as in the works of Brooks and Philip Agre\footnote{Philip E. Agre, \emph{Computation and Human Experience} (Cambridge University Press: 1997).} fell short of engaging with the world in a truly nonrepresentational way. Historically, then, we could understand object recognition as a simple perceptual appendage onto symbolic AI, even in its ``state of the art'' manifestation at the beginning of the 2020s.

In digital art history, the essentially anagraphic quality of object recognition sparked attempts to at least improve the ontologies of pre-trained systems. ``Fine-tuning'' such systems, for a while, promised to liberate computer vision from its problematic and anachronistic training datasets, or at least integrate art historical images. Some efforts focused on object recognition in non-photographic images (which computer scientists have called the ``cross-depiction problem''\footnote{Peter Hall, Hongping Cai, Qi Wu, and Tadeo Corradi, ``Cross-depiction Problem: Recognition and Synthesis of Photographs and Artwork,'' \emph{Computational Visual Media} 1 (2015): 91-103.}), others on recognising classes of object which were not present in ImageNet\footnote{C.f. the \emph{Saint George on a Bike} project of the Barcelona Supercomputing Center and Europeana, URL: \url{https://saintgeorgeonabike.eu/}}; but overall, fine-tuning failed to establish itself as a viable digital methodology for most digital art history projects. The reasons for this are many -- they surely include the massive data and compute resources necessary to retrain neural networks. Potential improvements were often insignificant; pretrained models (mostly trained on ImageNet) simply worked well enough for the limited (in the technical, not scholarly sense) tasks designed by digital art history researchers; whilst the pace of development in ``mainstream'' computer vision meant that by the time a model had been adapted for art-historical tasks, a better-performing model would have been published elsewhere.

It would also be untrue to say that object recognition models have no sense of the relationship between objects at all. Rather, their relationships are incidental and based on simple visual correlations. This can be seen in the mis-classifications made by object recognition systems. Amongst the 10 object classes of the CIFAR-10 dataset, for instance, one algorithm reports especially high confusions between the following pairs of classes: automobile-truck, dog-cat, deer-horse, and bird-airplane (Liu and Mukhopadhyay 2018). Yet the algorithm has no understanding of what it means to be a vehicle or a pet. These class confusions are entirely visual: birds and airplanes, for instance, are both often photographed against the sky. The object recognition model's internal representations mirror entirely visual relations between objects in the world.

More importantly, however, these class confusions suggest that there might be a more general utility to the internal representations of deep neural networks. Following the hypothesis that a successful classifier must have obtained at least some useful knowledge about the structure of
images (in the dataset), ``feature extraction'' uses selected internal representations to create ``embeddings'' of images, compressed representations of images as seen by the neural network. Unlike in word embedding models, where the distributional hypothesis suggests that word correlation has semantic implications at scale,\footnote{The classic example is the emergence of rudimentary analogic reasoning capabilities in word embedding models, such that analogy questions like ``what is to woman what king is to man'' can be answered by simple vector mathematics (``king + woman - man'').} for object recognition models such embeddings remain determined by visual features, and not by any extrinsic understanding of how objects relate to each other in the world. The ability to group together all images containing ``horse-like'' objects only gets you so far in addressing art-historical questions.

Ultimately, then, the use of embeddings extracted from object recognition models operates within the confines of a labeled visual world. In the context of Drucker's distinction between the digital and the merely digitized, we might consign object recognition to the latter. Though technically impressive, object recognition ultimately serves to facilitate access to images in an automated extension of traditional, categorical metadata. And yet: that it is possible to operationalize notions of similarity that go beyond what would be achievable on the pixel level even with object recognition models, points to the significant potential of image embeddings in particular. But to fully exploit this potential, we need to turn to a more recent class of models in which image embeddings can lead us \emph{beyond} the object recognition paradigm as they function as \emph{deep descriptions of semantically-contextualized visual concepts}: multimodal foundation models.

\section{Multimodal Foundation Models}\label{multimodal-foundation-models}

In multimodal foundation models, we are dealing with two interlocking technical developments. ``Foundation model'' is a term introduced by researchers at the Stanford HAI institute in 2021. It basically means: models that are a) very large, and b) that can be used for a variety of ``downstream'' tasks. ``Multimodality'' implies a conjoined processing of text and image data.

In the first instance, their multimodality allows them to encode both text and image data into a common space. They learn not from sets of images with discrete categories, but from pairs of images and texts -- digital images with descriptive captions. In practice, this means that their internal representational logic -- the shape of their embedding-space -- is informed by both visual and linguistic relations. Rather than being constrained by a fixed set of object categories (television, bird, airplane), multimodal models are able to encode an arbitrary textual input (up to a certain length) and images that are well-described by that text. These models are implicitly ekphrastic: they learn about texts through images, and vice versa; they necessarily structure both media in relation to the other. This means that they can describe things which are not \emph{objects}, and that they encode both visual and linguistic relations between them; mapping the relation between ``airplane'', ``car'', ``boat'', ``travel'', ``vehicle'' in both words and images.

The second distinction is one of scale. Foundation models, which include unimodal text models like GPT-3, are those that are trained on previously inconceivable amounts of data. ImageNet\footnote{ImageNet (in its reduced ILSVRC form) was the most popular research dataset of the pre-multimodal age and facilitated many early deep learning breakthroughs, see for instance Alex Krizhevsky, Ilya Sutskever, and Geoffrey E. Hinton, ``ImageNet Classification with Deep Convolutional Neural Networks,'' \emph{Advances in Neural Information Processing Systems} 25 (2012).} contains around 14 million images in roughly 22,000 categories; at time of writing, the most common datasets for training text-image models are 400 million and 5 billion (LAION-400M and LAION-5B) image-text pairs respectively. ImageNet is intended to be entirely made up of photographs, ideally with clear, isolated objects at their center; LAION contains drawings, paintings, and born-digital images.

Thus, multimodal foundation models allow for a surprisingly nuanced exploration of complex visual concepts which fall outside the conceptual constraints of object recognition. This affords a new set of practices when working with visual data. At the same time, it also creates new urgency for understanding the ``visual culture'' of such trained models, precisely because it enables us to look for visual concepts (``violence'', ``rhythm'', ``70s neighborhood'') that necessarily only exist in a culturally-situated way.

Historically, the paper ``Learning transferable visual models from natural language supervision''\footnote{Alec Radford, Jong Wook Kim, Chris Hallacy, Aditya Ramesh, Gabriel Goh, Sandhini Agarwal, Girish Sastry, et al., ``Learning Transferable Visual Models from Natural Language Supervision'' \emph{International Conference on Machine Learning (ICML)}, 2021: 8748--63.} was a turning point initiating the current paradigm shift. The paper proposes a multimodal foundation model architecture called ``CLIP'', with both architecture and a pre-trained
model released on GitHub. CLIP showed impressive capabilities on a wide range of tasks, including ``zero-shot'' image classification, i.e. the classification of images belonging to classes not specifically observed during training. However, as it was released at the same time as the first DALL-E model, a generative model with impressive image synthesis capabilities and much more discursive impact, CLIP's potential for the digital humanities was not realized until a few months later, when a handful of researchers started to use it as an embedding model in the way described above. CLIP's image embeddings, it turned out, not only surpassed those of previous models in ``quality'' but also facilitated a completely new form of image retrieval based on its multimodal capabilities. Early multimodal search engines like imgs.ai\footnote{Fabian Offert and Peter Bell, ``Imgs.Ai. A Deep Visual Search Engine for Digital Art History,'' \emph{International Journal for Digital Art History} (forthcoming).} offered an interface to this functionality.

\begin{figure}[h]
\centering
\includegraphics[width=\linewidth]{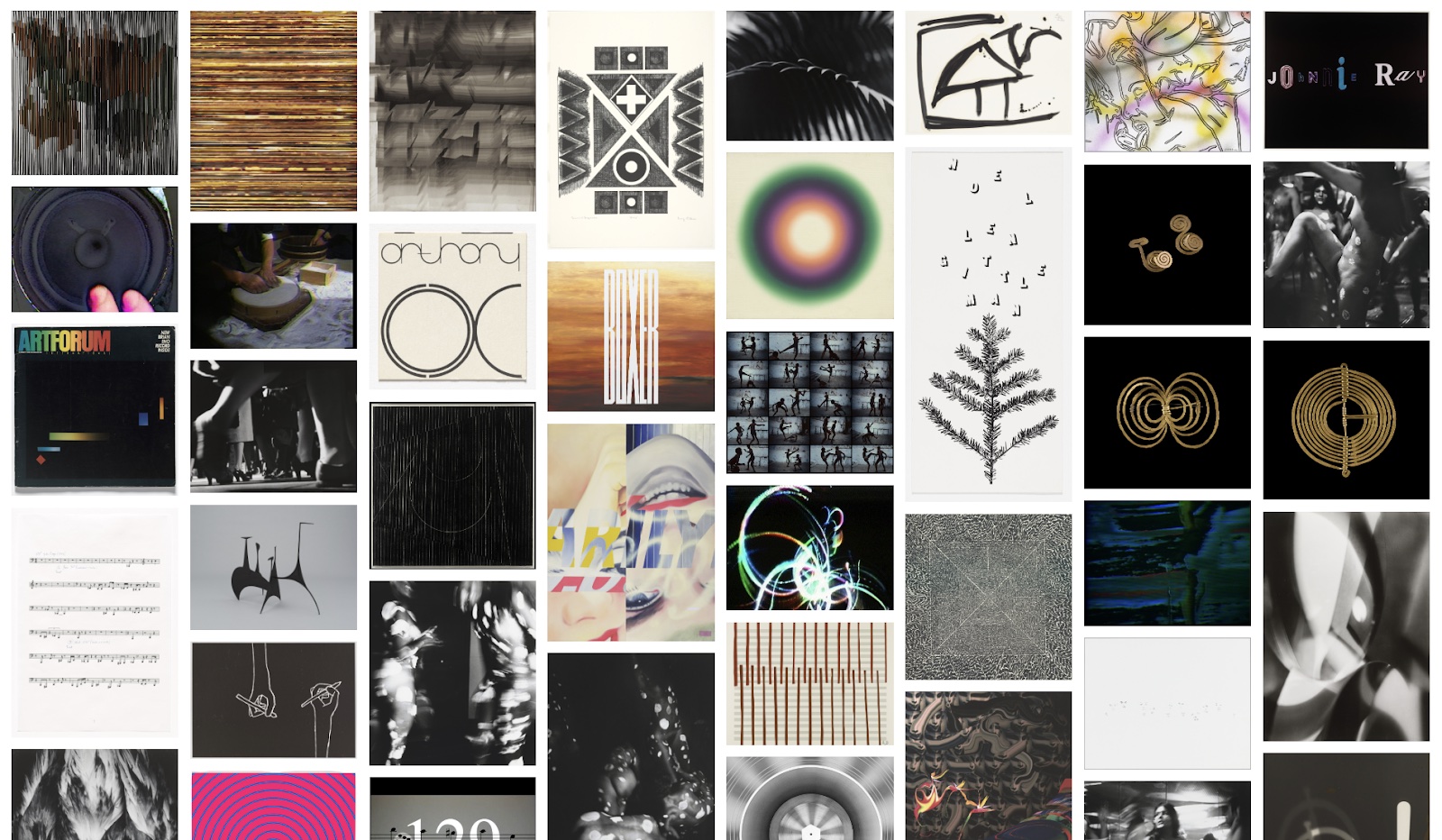}
\caption{Images for the keyword ``rhythm'' in the Museum of Modern Art, New York, collection, produced by the imgs.ai search engine. The results show the polyvalence of the knowledge embedded in the CLIP visual artificial intelligence model. Sheet music, album covers, photos of audio equipment, images containing the letters in the word ``rhythm'', images that could be interpreted as waveforms or spectrograms, as well as ``rhythmic'' graphical works are returned.}
\label{fig2}
\end{figure}

Through multimodal search it became possible to implement trivial but difficult to operationalize retrieval tasks like finding images of specific objects, e.g. images of chairs in the Museum of Modern Art, New York, collection, or exploring certain motifs in a collection of magic lantern slides.\footnote{Thomas Smits and Melvin Wevers, ``A Multimodal Turn in Digital Humanities. Using Contrastive Machine Learning Models to Explore, Enrich, and Analyze Digital Visual Historical Collections,'' \emph{Digital Scholarship in the Humanities,} 2023} CLIP, in other words, mirrors the state of the art of object recognition. But more importantly, other than in metadata-based retrieval systems and object recognition models, CLIP, through its natural language interface, allows for retrieval based on \emph{visual concepts of arbitrary complexity}. A CLIP search for ``summer'' will turn up summer-like images: of beaches, pools, people in swimsuits, or landscapes drenched in yellow light. A CLIP search for ``rhythm'' (figure \ref{fig2}) will retrieve images which embody a large spectrum of meaning inherent in that word: images of sheet music, album covers, and loudspeakers, works that resemble oscilloscope graphs or spectral plots, or graphical works that involve regular patterns that could be described as somehow rhythmic. We are still dealing here with image retrieval, but CLIP takes us far beyond the object recognition paradigm.

Let us consider another example that demonstrates these capabilities of the CLIP model. Diego Velázquez' 1656 painting ``Las Meninas'' is one of the most discussed pictures of art history. Michel Foucault, famously, spends the whole introduction of \emph{The Order of Things} on it. The painting is famous, in particular, for its play on representation. The painter himself is in the painting, but we do not see what he is painting, as we can only see the backside of the canvas --- or do we indeed see what he is painting, as we are potentially looking at the picture he is painting at that moment? There is a mirror which opens up another, invisible image space, and countless gaze relations tell an intricate story about the historical characters in the picture.

\begin{figure}[h!]
\centering
\includegraphics[width=\linewidth]{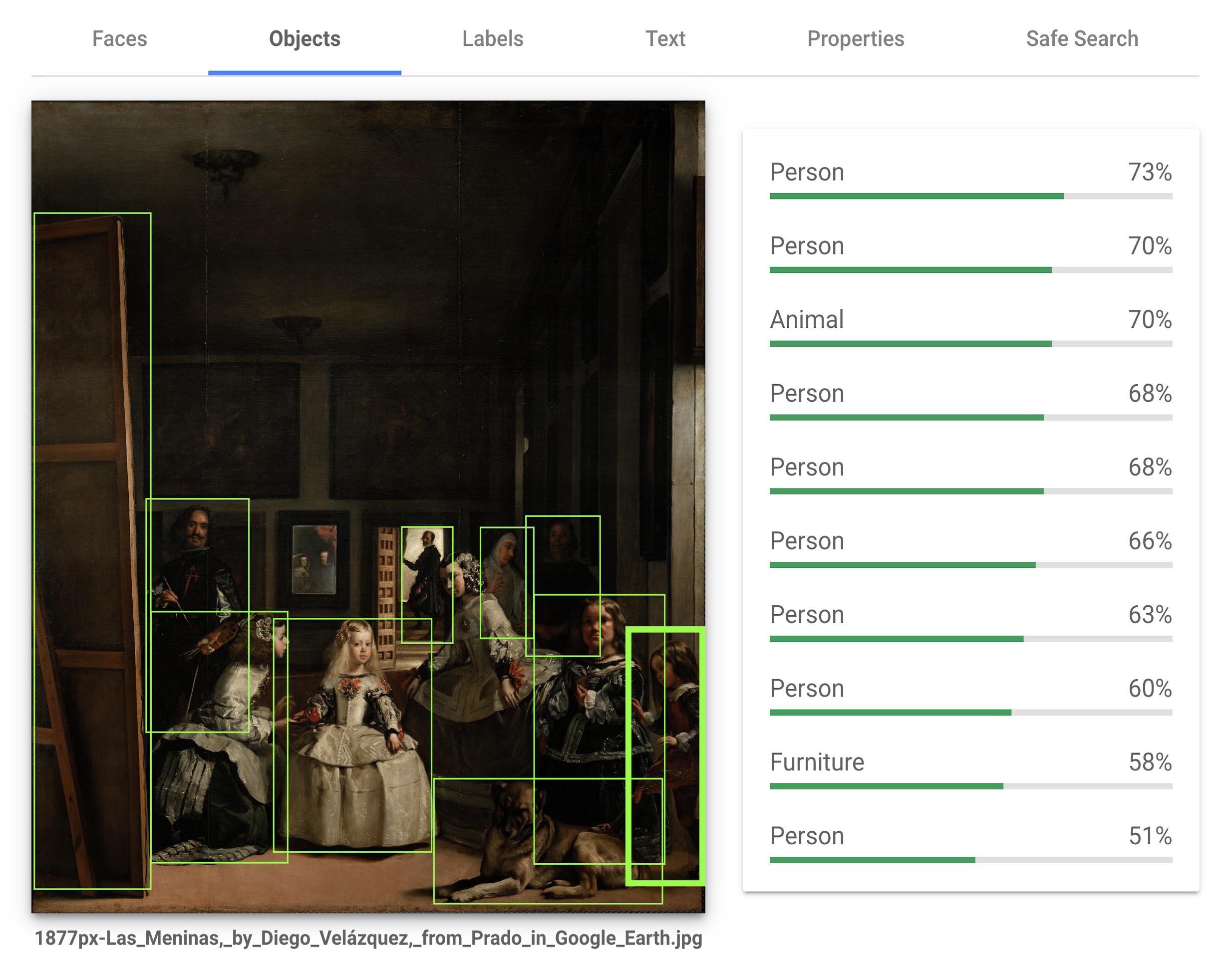}
\caption{Google cloud vision API results for \emph{Las Meninas}, easel is ``furniture''}
\label{fig3}
\end{figure}

Using the techniques of digital art history so far, what can we say about this picture? We might be able to determine the number of people in the picture with the help of a pre-trained and/or fine/tuned face detection network. We might confirm the existence of certain image objects --- an easel, a dog, other paintings with the help of an object detection network (figure \ref{fig3}). We might even be able to estimate the gaze direction of some of the characters in the picture. But under no circumstances could we infer the play on representation that the picture embodies, the fact that it is, with W.J.T. Mitchell, a ``metapicture'',\footnote{W. J. Thomas Mitchell, \emph{Picture Theory: Essays on Verbal and Visual Representation} (University of Chicago Press: 1995): 35.} a picture about pictures, a representation of representation.

If we run an imgs.ai search for the phrase ``Las Meninas'' on the collection of the Museum of Modern Art, New York, an institution that does not only not have the famous painting in its collection (which is kept in the Prado in Madrid) but also focuses on contemporary art in general, the results are surprisingly enlightening and show the conceptual depth that CLIP allows the user to access. Among them we find Richard Hamilton's \emph{Picasso's Meninas} from \emph{Homage to Picasso} (1973)\footnote{\url{https://www.moma.org/collection/works/60206}}, which takes up the structure of the Velásquez original but fills it with figures from Picasso paintings. While both works have nothing in common but their compositional structure, CLIP is still able to draw formal connections through their compositional similarity. CLIP also points us to two photographic works, Joel Meyerowitz' \emph{Untitled} from \emph{The French Portfolio} (1980)\footnote{\url{https://www.moma.org/collection/works/181924}} and Robert Doisneau's \emph{La Dame Indignée} (1948)\footnote{\url{https://www.moma.org/collection/works/52526}}; both explicit plays on representation, mirroring \emph{Las Meninas}' concern with the gaze relation between people in, and people before\footnote{Using George Didi-Huberman's term. See Georges Didi-Huberman, \emph{Confronting Images: Questioning the Ends of a Certain History of Art} (Penn State Press: 2005).} the image.

The question then is: what are the \emph{potential applications} and \emph{epistemic implications} of a system that has learned to ``understand'' -- or at least operationalize -- complex, para-visual concepts \emph{in terms of} visual attributes? How can it, in Drucker's terms, change the way we understand the objects of our inquiry?

What can we know about the para-visual concepts that CLIP operates on? While we can reason about them theoretically, or on the basis of our knowledge about training datasets, we want to suggest that it is precisely the emerging techniques of multimodal digital art history that can be used to investigate the shapes of CLIP's concepts, so to speak: and, by extension, their epistemic and ideological implications. Two experiments -- which are to be read as mere suggestions for potential directions -- will serve to demonstrate what we see as a necessary entanglement of object and model analysis. We make the tools developed for both experiments freely available on the Web\footnote{\url{https://leoimpett.github.io/clip-map/}, \url{https://leoimpett.github.io/2dclip/}} so that our results can be reproduced on demand.

\section{Conceptual Maps: What Makes Paris Look Like Paris?}\label{conceptual-maps-what-makes-paris-look-like-paris}

Again we will consider a concrete example: the city of Paris. CLIP has a concept, a mental image, of Paris. A CLIP-search for ``Paris'' in the MoMa collection, for instance, turns up images of the Tour Eiffel, of Notre Dame, and of generally ``French-looking'' streets. Paris, for CLIP, is bound up with a set of particular visual, we might say symbolic, properties; but how do they relate to the material reality of Paris as it exists today?\footnote{The inspiration for this experiment comes from a now classic computer science paper that aimed to isolate geographically salient architectural elements (``windows, balconies and street signs''), taking Paris as an example. See Carl Doersch, Saurabh Singh, Abhinav Gupta, Josef Sivic, and Alexei Efros, ``What Makes Paris Look Like Paris?'', ACM Transactions on Graphics 31 no. 4 (2021).} Where in Paris looks (to CLIP) like Paris -- and what does that tell us about CLIP?

\begin{figure}[h]
\begin{subfigure}[t]{.5\textwidth}
\centering
\includegraphics[width=\linewidth]{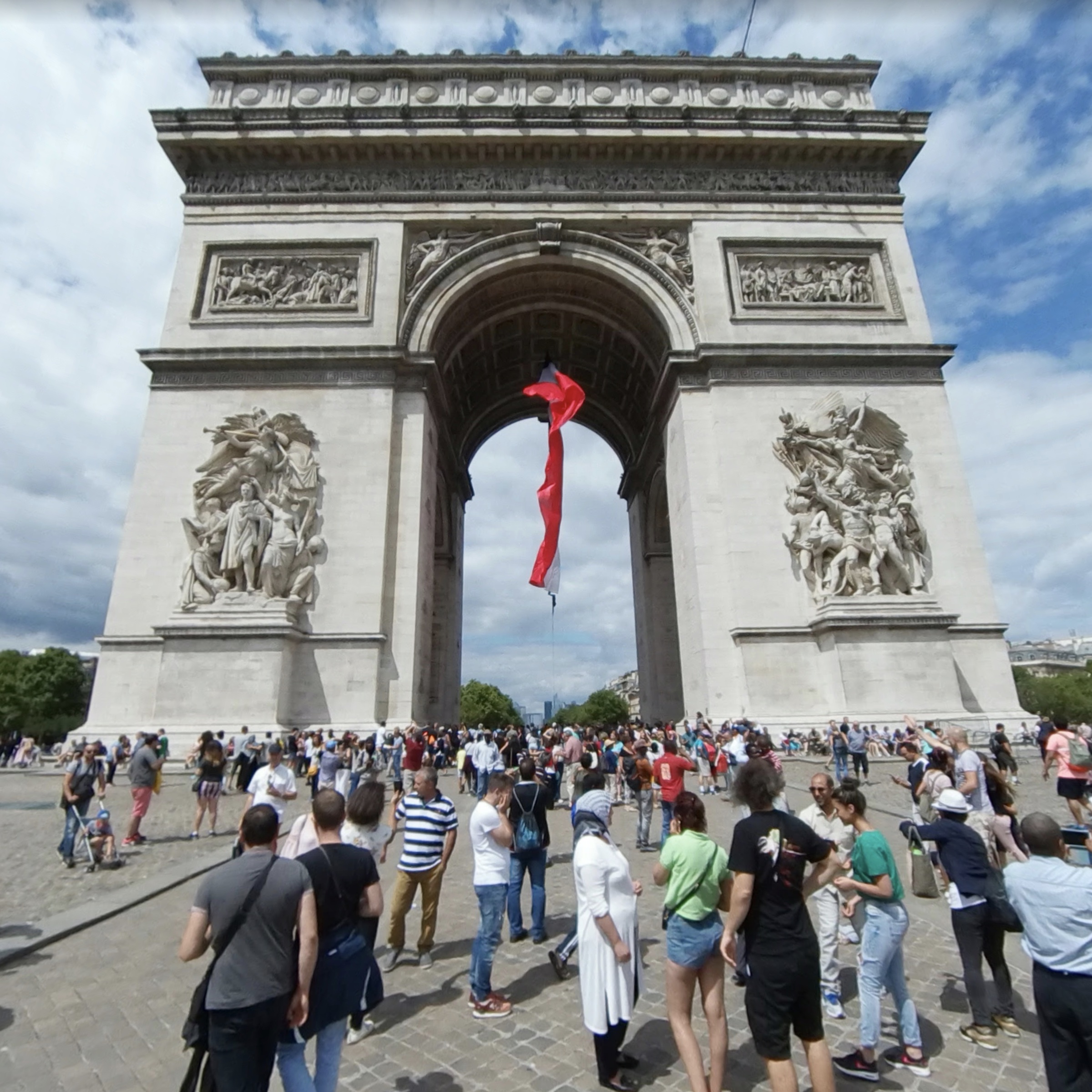}
\end{subfigure}
\begin{subfigure}[t]{.5\textwidth}
\centering
\includegraphics[width=\linewidth]{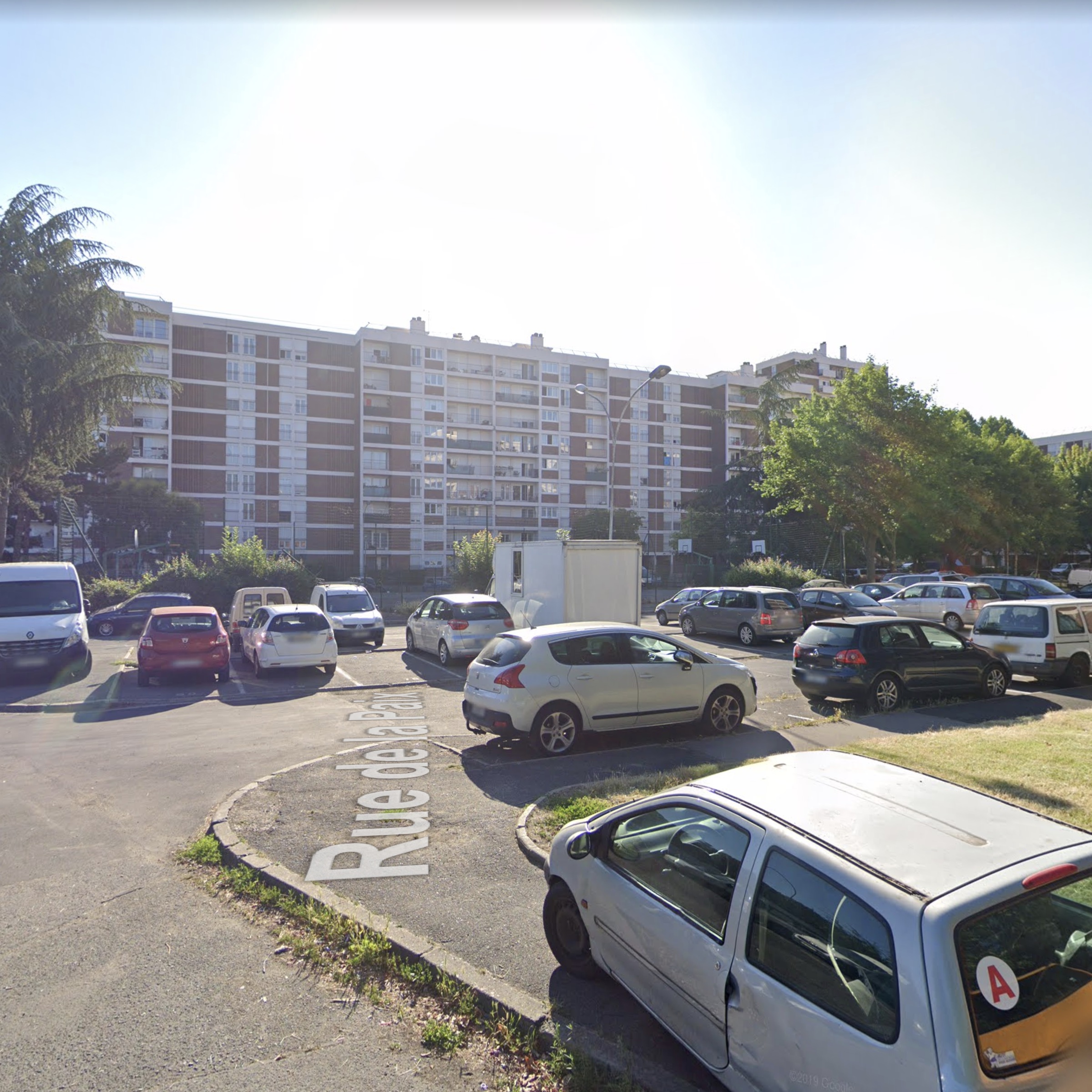}
\end{subfigure}
\begin{subfigure}[t]{.5\textwidth}
\centering
\includegraphics[width=\linewidth]{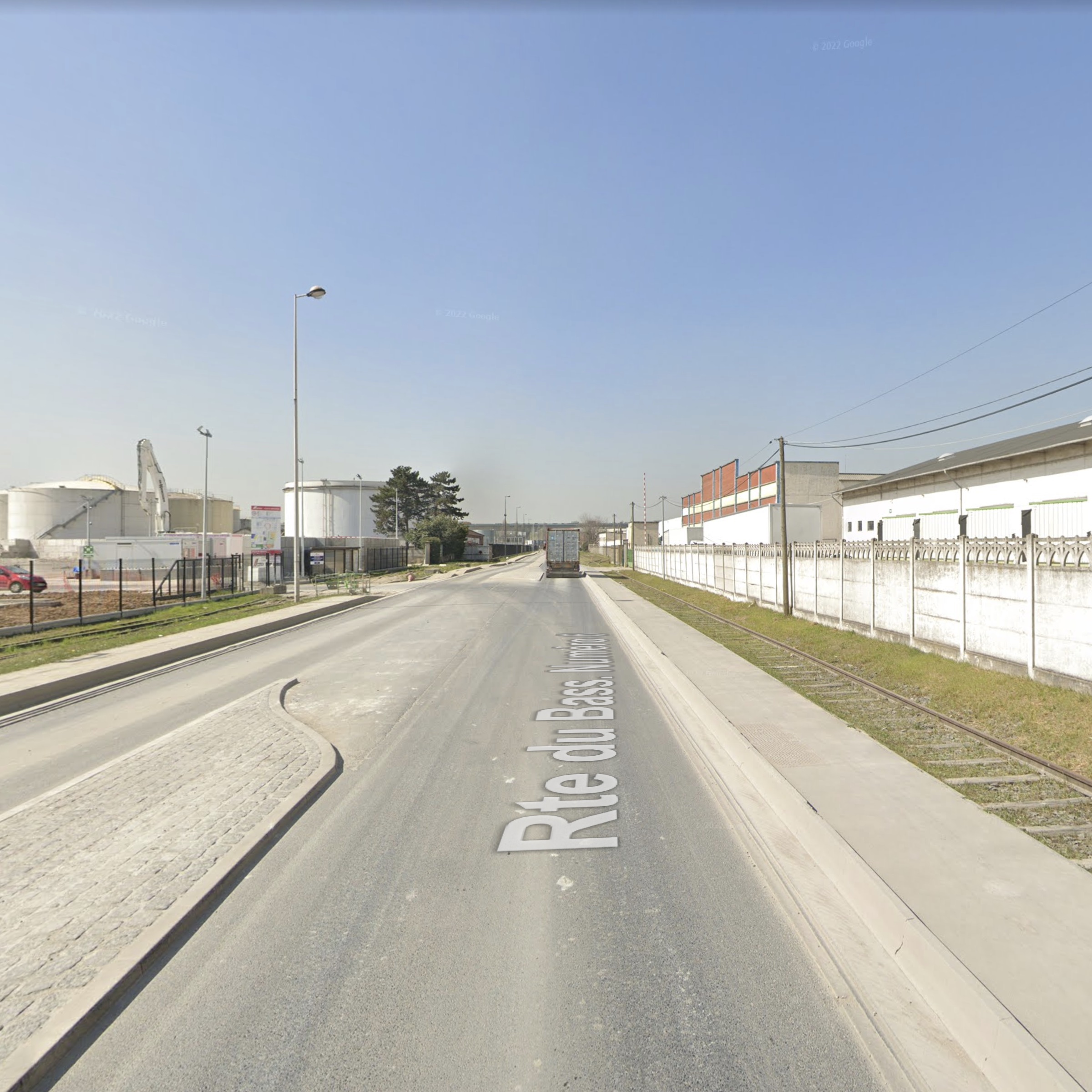}
\end{subfigure}
\begin{subfigure}[t]{.5\textwidth}
\centering
\includegraphics[width=\linewidth]{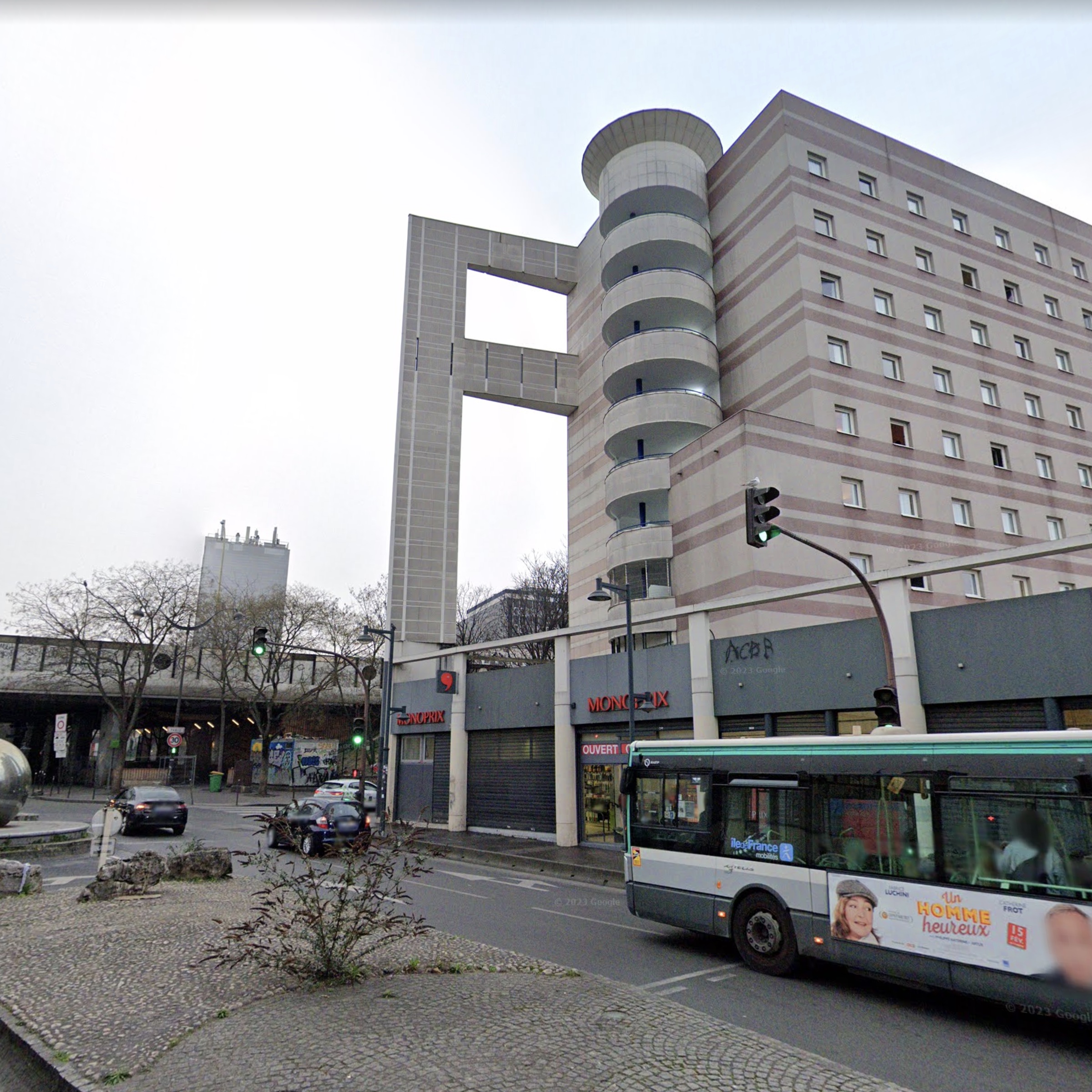}
\end{subfigure}
\caption{Left to right, top to bottom: two of the most and least Parisian images of Paris; a very ``Los Angeles'' image of Paris; the most ``New York'' image of Paris}
\label{fig5}
\end{figure}

\begin{figure}[h]
\begin{subfigure}[t]{.5\textwidth}
\centering
\includegraphics[width=\linewidth]{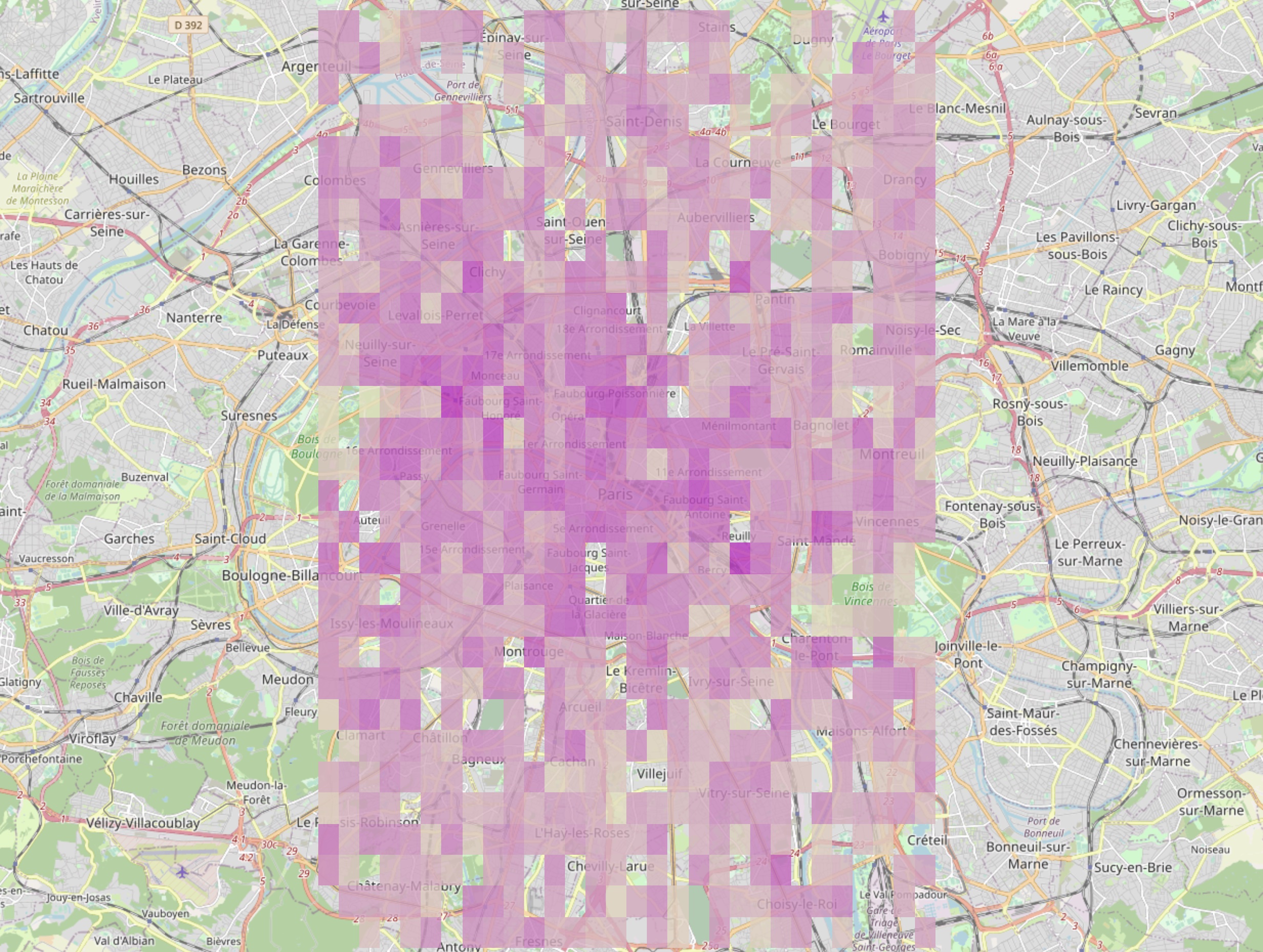}
\end{subfigure}
\begin{subfigure}[t]{.5\textwidth}
\centering
\includegraphics[width=\linewidth]{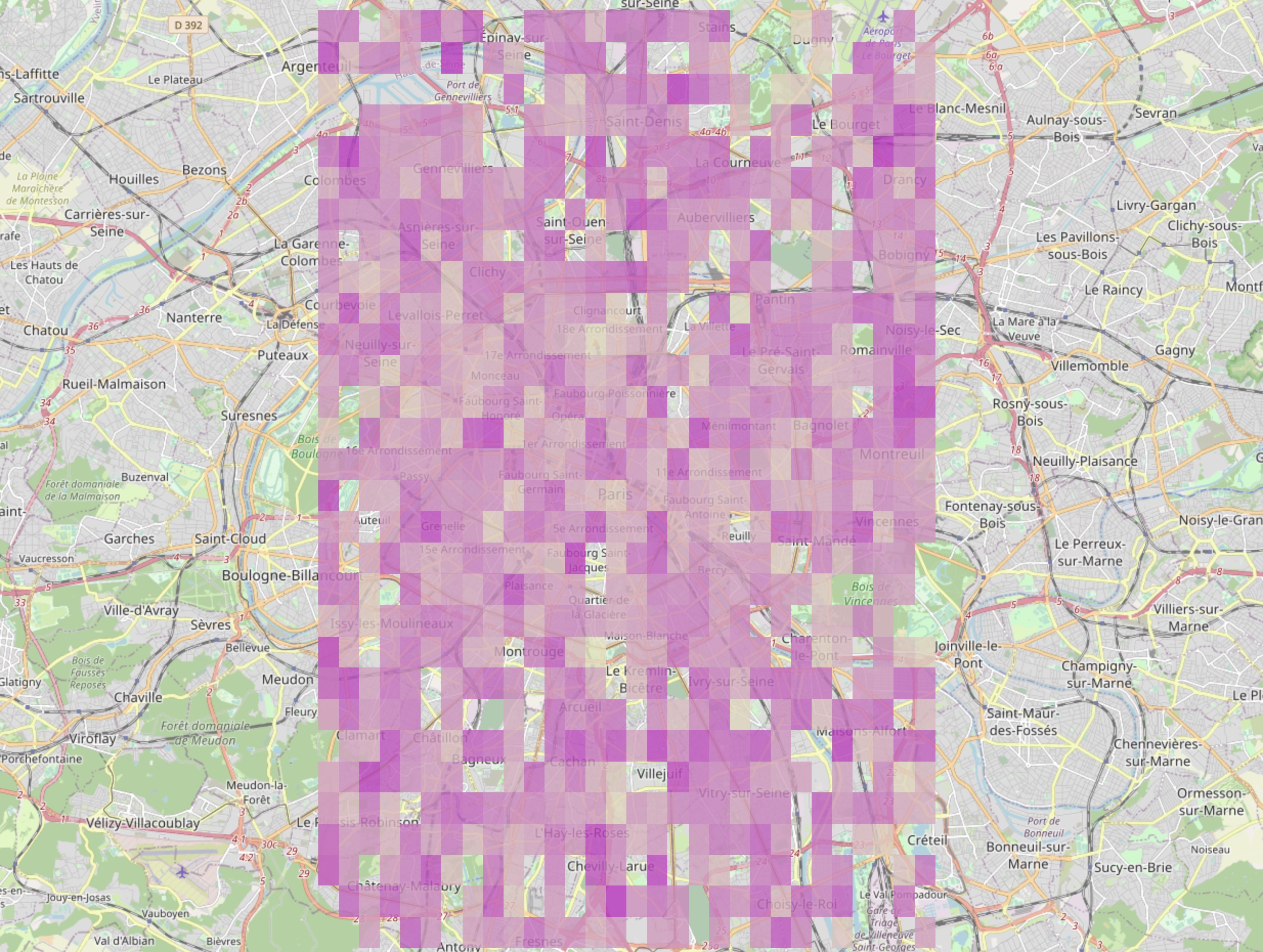}
\end{subfigure}
\caption{Strength of CLIP activations for sampled Google Street View images for ``a photo of Paris'' and ``a photo of Los Angeles''}
\label{fig6}
\end{figure}

To attempt to answer this question, we collect 10,000 images of Paris from Google Street View, sampled at regular intervals of latitude and longitude. All are images of Paris, the place; and yet not all are equally associated, in the visual logic of CLIP, with the prompt ``A photo of Paris''. If we plot the strength of the association of each image with that prompt in the CLIP latent space, in the form of a heat map overlay, we can see that CLIP's concept of Paris is highly spatially uneven (figures \ref{fig5} and \ref{fig6}). The city center produces the highest associations (especially landmarks like the Arc de Triomphe); as we get to the periphery, the score is significantly lower. CLIP's visual model, its symbolic imaginary, of \emph{Parisness} is one of landmarks and Hausmannian boulevards.

This is an extreme case, of course. And yet it points to a wider sense in which CLIP's model of anything is always a mental image, necessarily deeply culturally situated, linked to the image-economies of the internet: in this case, a visual imaginary of Paris linked both to tourism and to national identity (and state power). Every city (and more broadly, every visual concept) comes with its own highly-symbolic imaginary: the most ``Los Angeles'' parts of Paris, for CLIP, are to be found in the periphery, not in the center. One of the most ``dystopian'' parts of Paris is, at the same time, its most ``New York'' part -- the actually most ``dystopian'' image being an inside shot of a laser tag facility somewhere on the outskirts of the city.

Gardens, graffiti, public spaces, gentrification: one can imagine using multimodal networks such as CLIP in this way to literally \emph{map out} urban cultural geography; at the same time necessarily mapping the network's culturally-situated ``mental image'' of garden-ness, graffiti-ness, and so forth. Using CLIP as a research tool to study an image-corpus is always bound up with exploring CLIP's own way of seeing.

\section{2D CLIP: Naked and Nude}\label{d-clip-naked-and-nude}

Let's pivot our attention to the basic working unit of digital art history: the image corpus. Multimodal networks give us a new way of measuring visual phenomena: we can measure anything we can name across a dataset of images. But instead of using these text-image similarities as
rankings in a search engine or colors on a map, we can extract them as ``features''. Building on the work of Lev Manovich (amongst others), each feature can then become an axis on which to build a visualization of an image corpus.

One of the most powerful ways of probing (and indeed using) CLIP's mental images is to focus on the messy distinctions between a pair of highly entangled concepts, as with ``Paris'' the prompt, and ``Paris'' the set of street-view images. This is the intuition behind our browser-based visualization tool, \emph{2D CLIP}. In 2D CLIP, two separate CLIP-based visual concepts are measured \emph{against each other}: allowing the user to map an image corpus through, for instance, ``utopian'' versus ``science-fiction'', ``modernism'' versus ``modernity'', and so on.

\begin{figure}[h]
\centering
\includegraphics[width=\linewidth]{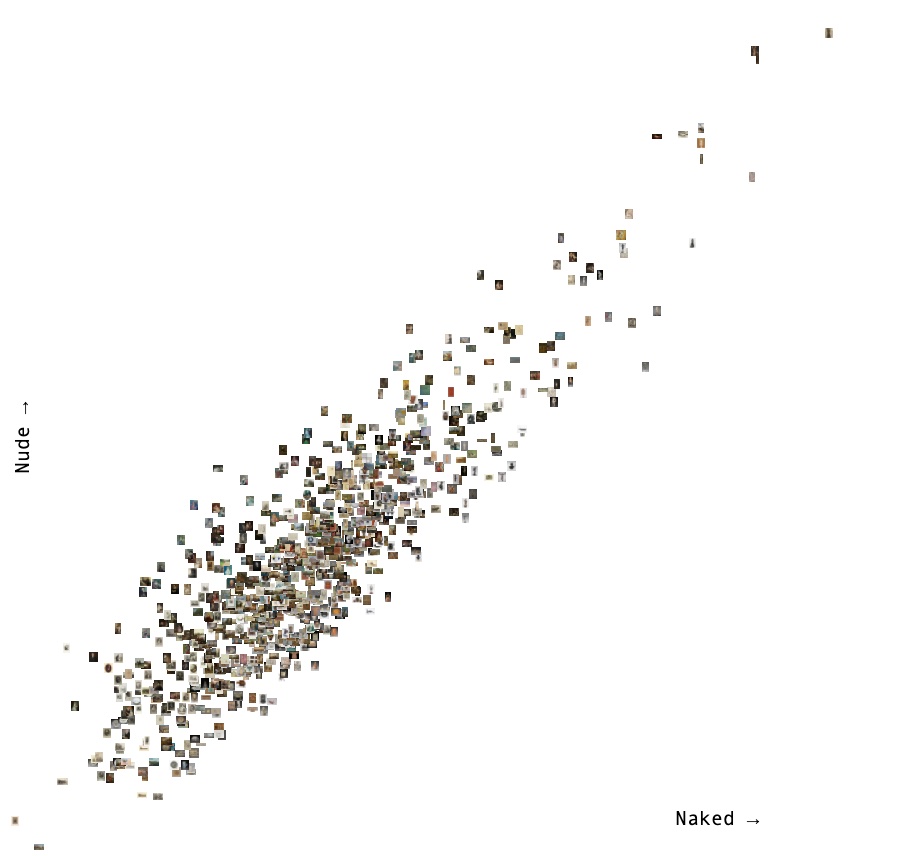}
\caption{2D CLIP interface with 1000 images from the Art Institute of Chicago collection scattered along two conceptual axes: ``naked'' and ``nude''}
\label{fig7}
\end{figure}

We can, for instance, map 1,000 images from the collection of the Art Institute of Chicago along their CLIP-similarities with the terms ``naked'' and ``nude'' (figure \ref{fig7}). The two concepts are so tightly bound up that most European languages do not distinguish between them, while at the same time they represent a historically relevant art-historical conceptual pair. Reacting to Kenneth Clark, for whom nakedness implied embarrassment and nudity a ``balanced, prosperous, and confident body'',\footnote{Kenneth Clark, \emph{The Nude. A Study in Ideal Form} (Princeton University Press: 1984): 3.} John Berger saw the nude in terms of the male gaze: a body can be naked on its own, but it ``has to be seen as an object in order to become a nude''.\footnote{John Berger, \emph{Ways of Seeing} (London, Penguin Books: 1972): 54.}

The images from the Art Institute follow a fairly straight diagonal line; images which are more naked, are also more nude. But there is some noise, some deviation to the pattern. Most images do not sit on the imaginary diagonal line, but just above (more nude) or below (more naked). For CLIP, Jules Joseph Lefebvre's late 19th century reclining \emph{Odalisque} (literally ``chambermaid'' but in the orientalist imaginary, closer to ``concubine'') is markedly more nude, as are a host of reclining and seated unclothed women. CLIP's nudes largely share this stillness, but there are exceptions: the sexual violence of Tintoretto's Tarquin and Lucretia, for instance. Ferdinand Hodler's \emph{Day (Truth)} is, by contrast, determinedly \emph{naked} for CLIP (fig. 8). How are we to read that against Clark's \emph{balanced, prosperous, confident}?

Again, this exercise does not just tell us about CLIP, but about the Art Institute's own collection (or our subsample of it). Most painted nudity is female nudity, of course. Unclothed male bodies are to be found either together with female ones (Adam and Eve, Tarquin and Lucretia), or in sculpture. CLIP sees these largely classical male sculptures as decidedly \emph{naked}. Strange, given their central place in the imaginary of the nude: they include a Roman marble Meleager and a bronze Hercules after Lysippos, whom Clark had said created ``some of the finest nudes in art''.\footnote{Clark, \emph{The Nude}: 181.} Though rendered somewhat androgynous by its missing head and limbs, even the Art Institute's Knidos Aphrodite -- a copy of Praxiteles, no less -- is seen by CLIP as less nude than naked.

The nude, for CLIP, is not only gendered but also dependent on medium. And the privileged medium -- following the logic of computer vision -- is photography, with Alfred Stieglitz's photograph of Georgia O'Keeffe's bare torso being seen by CLIP as both more naked, and more nude, than the rest of the collection: we find it at the extreme top-right of the distribution. The most naked image is the most nude: we might turn to Lynda Nead, who insists that the two are always inseparably entangled, that there ``can be no naked ``other'' to the nude, for the body is always already in representation.''\footnote{Lynda Nead, \emph{The Female Nude}, (Routledge: 2001): 16.}

\begin{figure}[h]
\begin{subfigure}[t]{.7\textwidth}
\centering
\includegraphics[width=\linewidth]{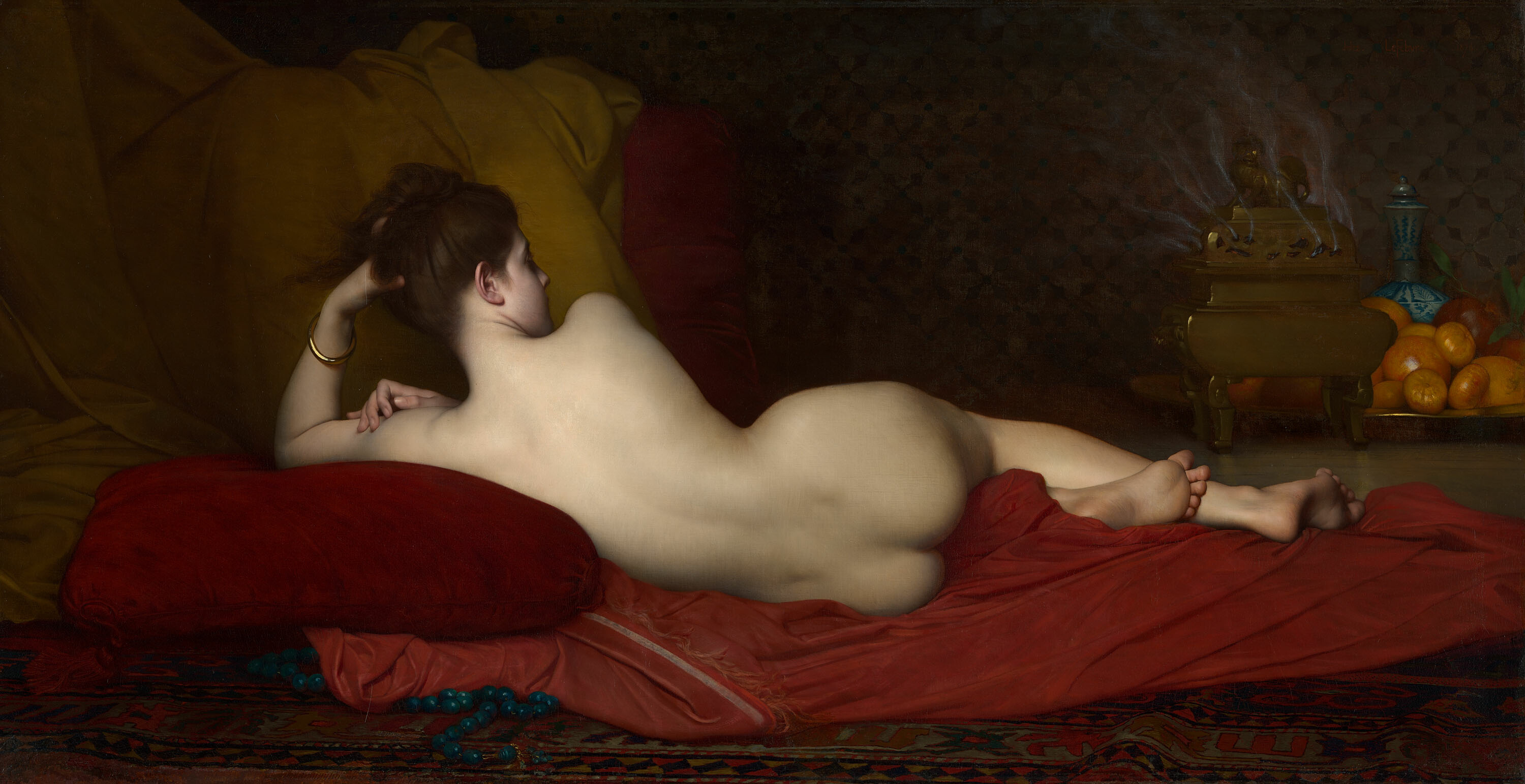}
\end{subfigure}
\begin{subfigure}[t]{.3\textwidth}
\centering
\includegraphics[width=\linewidth]{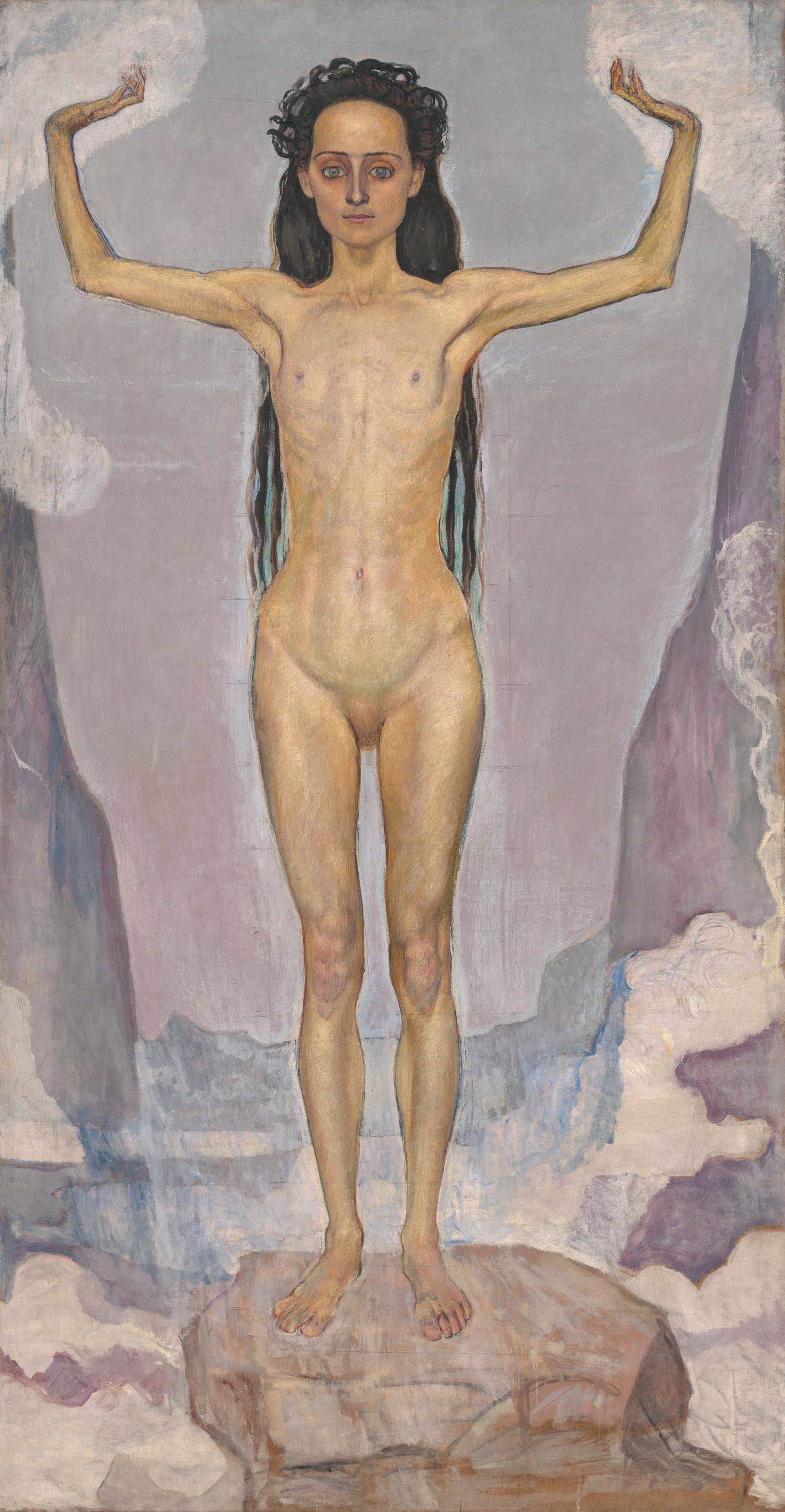}
\end{subfigure}
\caption{``Nude'' and ``Naked'' -- Jules Joseph Lefebvre's
\emph{Odalisque} (1874) and Ferdinand Hodler's \emph{Day (Truth)} (1896/98)}
\label{fig8}
\end{figure}

The operationalized naked-nude conceptual entanglement -- which in itself requires much more interpretative work -- is just one of many potential applications of the tool we developed for the purpose of this analysis, called 2D CLIP. Other than all remaining utilizations of CLIP in an art historical context, the tool runs in the browser but client-side, and is thus independent of network limitations and server-side restrictions. It allows researchers to load their own, locally hosted image collections into an accessible interface to create their own conceptual scatter plots.

\section{Conclusion}\label{conclusion}

We have deliberately covered a wide range of potential areas of interest in this text, ranging from spatial analysis to art-historical close-reading. Many more case studies could be facilitated by the 2D CLIP and CLIP MAP tools, or the general approach behind those tools. Our main contribution, however, is methodological.

Before the age of foundation models, neural networks never had an interesting model of human visual culture. The many deficiencies of ImageNet and its cousins were so easily exposed (for instance in the work of Travor Paglen or Adam Harvey) because of its claim to universality; a claim which was already made for the simple conceptual ontology ImageNet builds upon.\footnote{ImageNet builds upon the ontology of WordNet, see \hspace{0pt}\hspace{0pt}George A. Miller, Richard Beckwith, Christiane Fellbaum, Derek Gross, and Katherine J. Miller, ``Introduction to WordNet: An On-line Lexical Database,'' \emph{International Journal of Lexicography} 3 no. 4, 1990: 235-244.} Today, the makers of Stable Diffusion announce that their model ``is the culmination of many hours of collective effort to create a single file that compresses the visual information of humanity into a few gigabytes''.\footnote{Emad Mostaque, ``Stable Diffusion Public Release,'' URL: \url{https://stability.ai/blog/stable-diffusion-public-release} (accessed 7/15/2023).} This is still a ridiculous thing to say, of course: digitized images are nothing but the tip of the vast iceberg of human image production. And yet, while contemporary visual models are often still not good enough to facilitate those hard technical challenges that artificial intelligence researchers are actually aiming for (like detecting cancer or driving a car), they contain so much more information about human visual culture\footnote{The information learned by such models has, of course, still to be contextualized in relation to potentially significant dataset and inductive biases. For the former, see Abeba Birhane, Vinay Uday Prabhu, and Emmanuel Kahembwe, ``Multimodal Datasets: Misogyny, Pornography, and Malignant Stereotypes,'' arXiv preprint 2110.01963 (2021) or the many resources at \url{https://knowingmachines.org} (accessed 7/15/2023). For the latter, see Eva Cetinic, ``The Myth of Culturally Agnostic AI Models, arXiv preprint 2211.15271 (2022), and Fabian Offert, ``On the Concept of History (in Foundation Models),'' \emph{IMAGE} 37, 2023. As Ted Underwood writes: ``approaching neural models as models of culture rather than intelligence gives us more reason to worry about them. But it also gives us more reason to hope.'' Ted Underwood, ``Mapping the Latent Spaces of Culture,'', \emph{Startwords} 3, August 2022, URL: \url{https://startwords.cdh.princeton.edu/issues/3/mapping-latent-spaces/} (accessed 7/15/2023). See also Sonja Drimmer's and Christopher J. Nygren's recent ``axioms'' on art history and artificial intelligence, where they caution against an uncritical adoption of artificial intelligence in art history in general, and digital art history in particular: Sonja Drimmer and Christopher J. Nygren, ``Art History and AI: Ten Axioms'', \emph{International Journal for Digital Art History} 10, 2023.} that digital art history might be the one discipline where they actually will have an outsized impact. But their usefulness -- this is the main argument we would like to make with this text -- hinges on a reevaluation of what digital art history actually is as a field.

To put it bluntly: we wanted digital art history, and we got digital art history, but it is not what we expected. If we want to go beyond the object recognition paradigm, if we want to go beyond merely taking stock of images and image objects, if we want to integrate close and distant viewing, if, in other words, we want to move towards a ``digital'' art history, we have to accept that the scope of the field needs to expand. Models -- and their idiosyncratic ways of seeing the world -- are our responsibility now, and any art-historical study harnessing the power of contemporary machine learning must necessarily, at least in part, also be a study \emph{of} contemporary machine learning.\footnote{As Fabian Offert and Peter Bell put it, ``critical machine vision'' turns out to be ``a humanities challenge''. Fabian Offert and Peter Bell, ``Perceptual Bias and Technical Metapictures. Critical Machine Vision as a Humanities Challenge,'' \emph{AI \& Society} 36, 2021: 1133--44. See also David Berry, ``The Explainability Turn,'' \emph{Digital Scholarship in the Humanities}, 2023.}

To translate this into disciplinary terms: digital art history needs to open up to influences, and cross-collaborations from two directions. First, from media studies. If we truly cannot have our cake and eat it too -- if we cannot harvest the utility of foundation models without them becoming our object of study -- we need a critical apparatus that is up to the task, a way to talk about visual models and models of the visual. Recent work in critical AI studies\footnote{See for instance the recent \emph{American Literature} special issue on ``Critical AI: A Field in Formation'', edited by Rita Raley and Jennifer Rhee (\emph{American Literature} 95 no. 2, 2023).} in particular -- a field which itself is only forming now -- suggests that this methodological revision is already on the way in other disciplines. Second from computational literary studies. Scholars like Ted Underwood, Rita Raley, or Matt Kirschenbaum have been considering the impact of large language models on the study of literature since the early days of the BERT model. Humanist critique has always emphasized the split between the (formal) language of computers and their perceived affordances, e.g. an ideologically discrete view of the world where everything is a binary distinction, and the potential for ambiguity and polysemy in natural languages. In large-scale vision models, however, the medium of formal inference is now natural language --- or, at the very least, a vector space which appears to be equally continuous, ambiguous, and polysemic. Art history is used to negotiating the mediation of the verbal; but we now triangulate between the double mediation of text and code. This complicates the critical analysis of machine learning systems in general, but also presents an opportunity to consolidate humanist and technical critique. ``Prompt engineering'' means employing the complex, multilayered features of language that humanists already have at their disposal as tools of visual analysis.

The discipline of art history, we conclude, is well prepared for this shift. In fact, we would like to suggest that art history is uniquely well prepared for what is to come. Horst Bredekamp, in the early 2000s, argued that the ``split'' of visual studies from art history was artificial, and that at its core it is a mere disciplinary issue . Likewise, the tasks ahead of digital art history fit into what art history does best: historically and analytically situating complex objects of visual culture.

\end{document}